\documentclass[conference]{IEEEtran}
\IEEEoverridecommandlockouts
\usepackage{cite}
\usepackage{amsmath,amssymb,amsfonts}
\usepackage{algorithmic}
\usepackage{graphicx}
\usepackage{textcomp}
\usepackage{xcolor}
\def\BibTeX{{\rm B\kern-.05em{\sc i\kern-.025em b}\kern-.08em
    T\kern-.1667em\lower.7ex\hbox{E}\kern-.125emX}}
\begin{document}

\title{Fine Tuning With Abnormal Examples
}

\author{\IEEEauthorblockN{Will Rieger, \textit{The University of Texas - Austin}}
\IEEEauthorblockA{\textit{wrieger@utexas.edu}}
}

\maketitle

\begin{abstract}
Given the prevalence of crowd sourced labor in creating Natural Language processing datasets, these aforementioned sets have become increasingly large. For instance, the SQUAD dataset currently sits at over 80,000 records. However, because the English language is rather repetitive in structure, the distribution of word frequencies in the SQUAD dataset's contexts are relatively unchanged. By measuring each sentences distance from the co-variate distance of frequencies of all sentences in the dataset, we identify 10,500 examples that create a more uniform distribution for training. While fine-tuning ELECTRA [4] on this subset of examples reaches better performance to a model trained on all 87,000 examples. Herein we introduce a methodology for systematically pruning datasets for fine tuning reaching better out of sample performance.
\end{abstract}

\section{Introduction}
Whether it is because of extensive crowd sourced labor, better computing, or enhanced storage, there is a plethora of data available to researchers training Natural Language Processing models for question answering tasks. However, because most of these question answering datasets (specifically SQUAD) rely of encyclopedic contexts, the distribution of word choice is similar amongst most of the examples. Given that most question answering tasks begin by predicting the starting or ending location where the answer might be within the context. Intuitively, it is reasonable to believe that by using a relative location in the context and some dataset artifacts this value can be reached using a set of non-linear models.

While, obviously, there has been great success in recent years using these kinds of models for this task, relative performance has increased somewhat asymptotically to that of human performance. Our hypothesis is that this ceiling has been created based upon underlying distributions in the data. No matter how large a training set becomes, it is not sufficiently robust until there is a sufficient number of examples that are mutually abnormal. If new examples added to the training set follow a similar lexical distribution artifacts will persist and not lead to models that are sufficiently robust.

Work has been done to similarly understand how models treat various examples through Dataset Cartography \cite{dataset_cartography}. Swayamdipta, et al. primarily focus on during training performance and variability. By simultaneously measuring the confidence and variability they categorize examples into three categories: easy-to-learn, ambiguous, and hard-to-learn. The methodology presented herein produces a similar grouping of examples. Our methodology produces three categories of examples: low abnormality, mutually abnormal, and high abnormality. We further go on to show that solely training on a subset of the most representative of those samples is sufficient for reasonably robust training.

\section{Measuring Abnormality}

Measuring this abnormality can be done through Mahalanobis Distance \cite{maha}. Developed in 1927, the Mahalanobis Distance was used to analyze distinct sub-groups in the features of human skulls. By measuring the relative distance from a multinomial distribution, a scalar value can be assigned to each sample and grouped accordingly. In practice, this score can be thought of as its relative abnormality to the distribution across these dimensions.

\begin{equation}
    d_t = (x_t - \mu) \Sigma^{-1} (x_t - \mu)'
    \label{eq:eq}
\end{equation}

For each example at step t, the Mahalanobis distance can be calculated by de-meaning the features of the example and multiplying it with the inverse of the Covariance matrix of the features of the entire dataset (Equation \ref{eq:eq}). For each of the 87,000 examples in the SQUAD dataset, the features of each context are defined as the density of each individual word's occurrence in the overall dataset. Additionally to properly compute the Covariance matrix, each context is back padded with 0's to ensure each example is the same length. 

\begin{figure}
    \centering
    \includegraphics[scale=0.35]{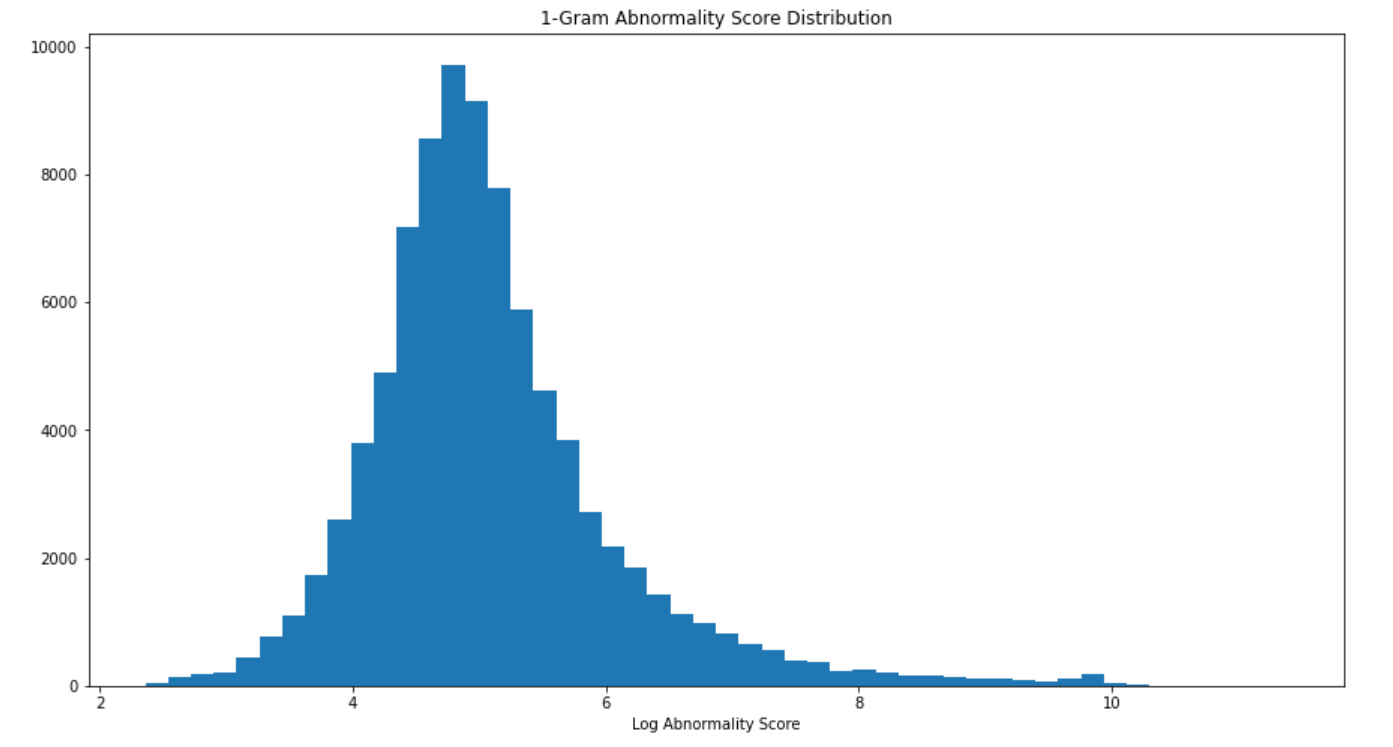}
    \caption{ Visualization of 1-Gram individual word abnormality. }
    \label{fig:1gram}
\end{figure}

Using each example's abnormality, we were able to examine the relative density of mutual abnormality between each example in the training set and the distribution of all examples' features (Figure \ref{fig:1gram}). The leptokurtic distribution of abnormalities supports our hypothesis that most examples with similar abnormalities to one another in their relative word-distribution. The fat tails of the low abnormality (left skew) and high abnormality (right skew) define sets of further abnormal examples better representing mutually abnormal examples (from one another). 

\section{Building the New Dataset}
In order to build a training set we sample 10,500 examples from each of the three categories defined. The high abnormality set is done by taking the indexes of the highest 3500 abnormality scores. The low abnormality set is done by taking the indexes of the lowest 3500 abnormality scores. The mutually abnormal set is done by sampling the 3500 indexes of examples closest to the mean of the distribution.

To better exemplify the differences between these three categories of examples, it is important to understand the categories of the contexts from each of the sampled distributions. For repeat-ability, the lowest abnormality score's index was example \#21706. The highest abnormality score's index was example \#12171. And the most mutually abnormal example has index \#9879. Anecdotally, the lowest example has the title "Brain", the highest example has the title "Space\_Race", and the mutually abnormal example has the title "Institute\_of\_technology".

\section{Training}
Following the aggregation of the 10500 training examples, the ELECTRA based model was trained for 3 epochs on the set. Each training loop took roughly 12 hours to complete. The complete training of ELECTRA took about 10 hours done on a Google Colab Pro GPU. So there is a noticeable speed increase by decreasing the size of the training set from 87,000 to 10,500 examples.

\section{Results}
Our initial ELECTRA model reached an F1 score on SQUAD of ~70.24 trained over 10 hours. The modified dataset reached an F1 score of 80.15 trained over 12 hours. We believe this notable increase in performance is due to better sampling the tails of the density distribution. Further we were able to naturally train the model with adversarial example by systematically choosing examples from the tails of the 1-gram distribution.

\section{Further Analysis}
\begin{figure}
    \centering
    \includegraphics[scale=0.35]{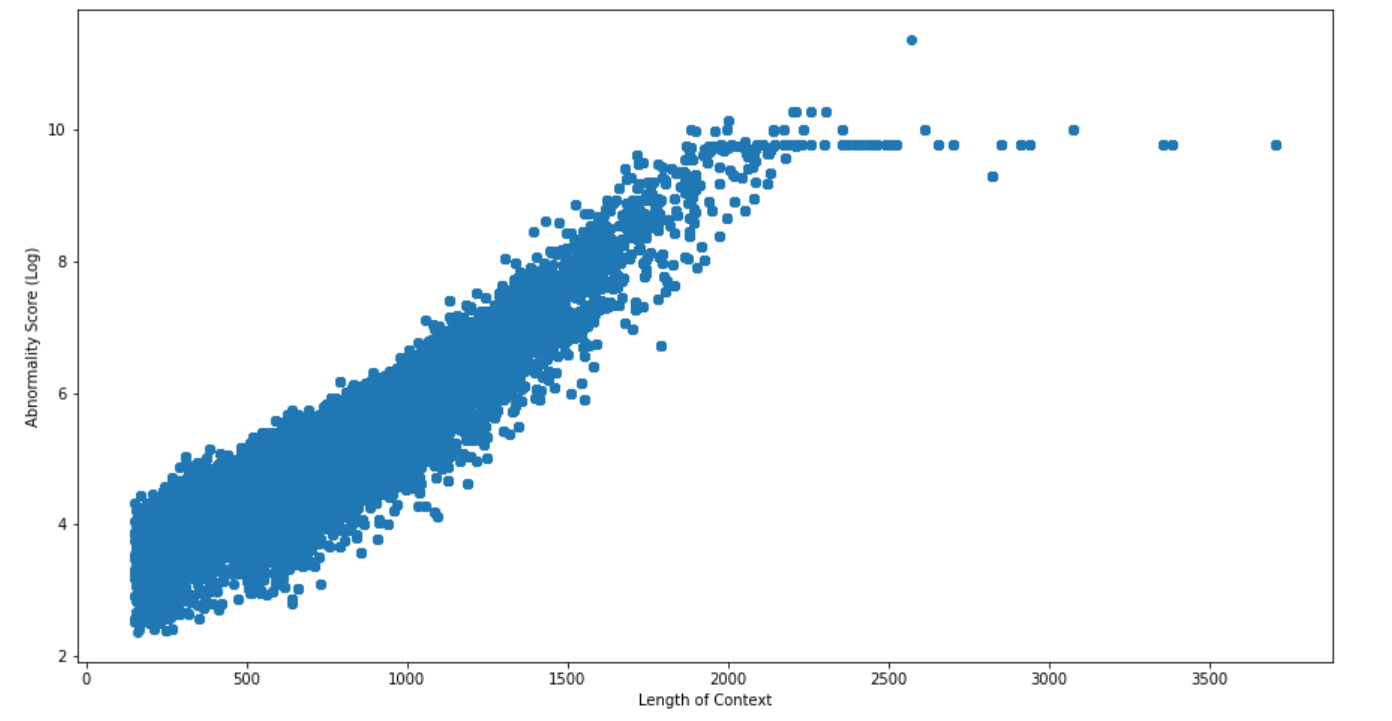}
    \caption{ Visualization of Context Length and 1-Gram Abnormality }
    \label{fig:1gram_correl}
\end{figure}

Upon further review our of our results, we noticed an interesting (certainly unintended) artifact in our 1-gram examples. There exists a linear relationship between the length of the context (in characters) and its abnormality score (Figure \ref{fig:1gram_correl}). While there is still a substantial variability in score, we believe that this unintended factor inclusion degraded our overall performance. Further investigating this phenomena we find that increasing the n-gram to 3, yields a less linearly correlated relationship between context length abnormality score. We reached the conclusion that individual word distributions would contain higher skew and therefore be more impacted by the matrix sparseness introduced by padding the shorter contexts. By increasing the length of the n-gram, the examples became more uniformly distributed and therefore less impacted by the sparseness impacted by padding.

During this review, we identified but were unable to test a way of using this revelation in future research. Ideally, in 250 character length segments of context, you sample from the high, low and mean of the distributions created to generate a training set. This would continue to support the initial hypothesis of dataset concentricity in the multiple dimensions of the lexical distribution. Additionally, this would yield a new sample distribution significantly less correlated to the dataset factor of length. In future work we plan to enhance the sampling methodology by sectioning abnormality within different baskets of length to remove the perceived correlation between length and abnormality.

\begin{figure}
    \centering
    \includegraphics[scale=0.35]{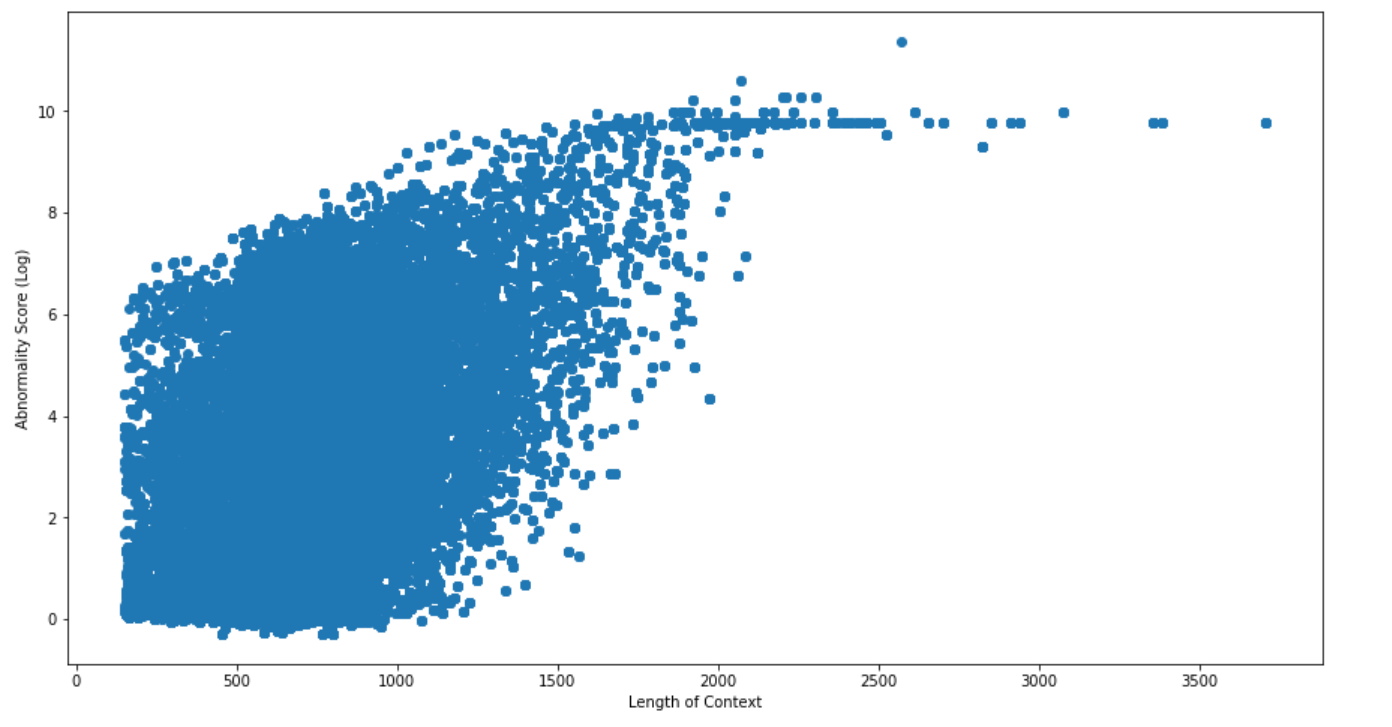}
    \caption{ Visualization of Context Length and 3-Gram Abnormality }
    \label{fig:3gram_correl}
\end{figure} 

\section{Conclusion}

We set out to investigate the relationship of lexical distribution and dataset crowding due to repetitiveness in the English language; especially related to encyclopedic datasets like SQUAD. We believe we made meaningful strides in the generation of a new methodology in defining the mutually abnormality of examples in a dataset versus this distribution. Future research is ongoing in creating subsets of data that meet the performance of training on the whole training dataset. The hope is that by exploring the avenues discussed in the section on Further Analysis these empirical artifacts (or features identified) may yield similarly or more robust performance in shorter fine-tuning training time.


\begin{thebibliography}{00}
\bibitem{maha} Mahalanobis, P.C. ``Analysis of Race-Mixture in Bengal`` in Journal of the Asiatic Society of Bengal, vol23:301-333, 1927.
\bibitem{squad-dataset} Rajpurkar, et. al. ``SQuAD: 100,000+ Questions for Machine Comprehension of Text`` in arxiv:1606.05250.
\bibitem{dataset_cartography} Swayambdipta, et. al. ``Dataset Cartography: Mapping and Diagnosising Datasets with Training Dynamics`` in arxiv:2009.10795v2.
\bibitem{electra-model} Clark, et. al. ``ELECTRA: Pre-training Text Encoders as Discriminators Rather Than Generators`` in arxiv:2003.10555.


\end{thebibliography}
\end{document}